\documentclass[letterpaper, 10 pt, conference]{ieeeconf}  

\IEEEoverridecommandlockouts                              
\usepackage[english]{babel}
\usepackage{amsmath}
\usepackage{amssymb}
\usepackage{multirow}
\usepackage{rotating}
\usepackage{multirow}
\usepackage{mathtools}
\usepackage{units}
\usepackage{subfloat}
\usepackage{float}
\usepackage[]{units}
\usepackage{xcolor}
\usepackage{todonotes}
\usepackage{subfig}
\usepackage{siunitx}
\usepackage{algorithm}
\usepackage{algorithmic}
\usepackage{footnote}
\usepackage{enumerate}
\makesavenoteenv{table}
\makesavenoteenv{table*}
\makesavenoteenv{tabular}
\makesavenoteenv{figure}
\usepackage{url}

\title{\LARGE \bf
Cable-Driven Actuation for Highly Dynamic Robotic Systems
}

\author{Jemin Hwangbo, Vassilios Tsounis, Hendrik Kolvenbach
 and Marco Hutter
\thanks{*This research was supported by the Swiss National Science Foundation through the National Centre of Competence in Research Robotics.}
\thanks{All authors are associated with Robotic Systems Lab at ETH Zurich, Switzerland.}
\thanks{$^{1}$\textit{jhwangbo@ethz.ch}}%
}

\begin{document}

\maketitle
\thispagestyle{empty}
\pagestyle{empty}

\begin{abstract}

This paper presents design and experimental evaluations of an articulated robotic limb called Capler-Leg. The key element of Capler-Leg is its single-stage cable-pulley transmission combined with a high-gap radius motor. Our cable-pulley system is designed to be as light-weight as possible and to additionally serve as the primary cooling element, thus significantly increasing the power density and efficiency of the overall system. The total weight of active elements on the leg, i.e. the stators and the rotors, contribute more than 60\% of the total leg weight, which is an order of magnitude higher than most existing robots. The resulting robotic leg has low inertia, high torque transparency, low manufacturing cost, no backlash, and a low number of parts. Capler-Leg system itself, serves as an experimental setup for evaluating the proposed cable-pulley design in terms of robustness and efficiency. A continuous jump experiment shows a remarkable \unit[96.5]{\%} recuperation rate, measured at the battery output. This means that almost all the mechanical energy output used during push-off returned back to the battery during touch-down.

\end{abstract}

\section{INTRODUCTION}
\begin{figure}
\centering
\includegraphics[width=0.43\textwidth]{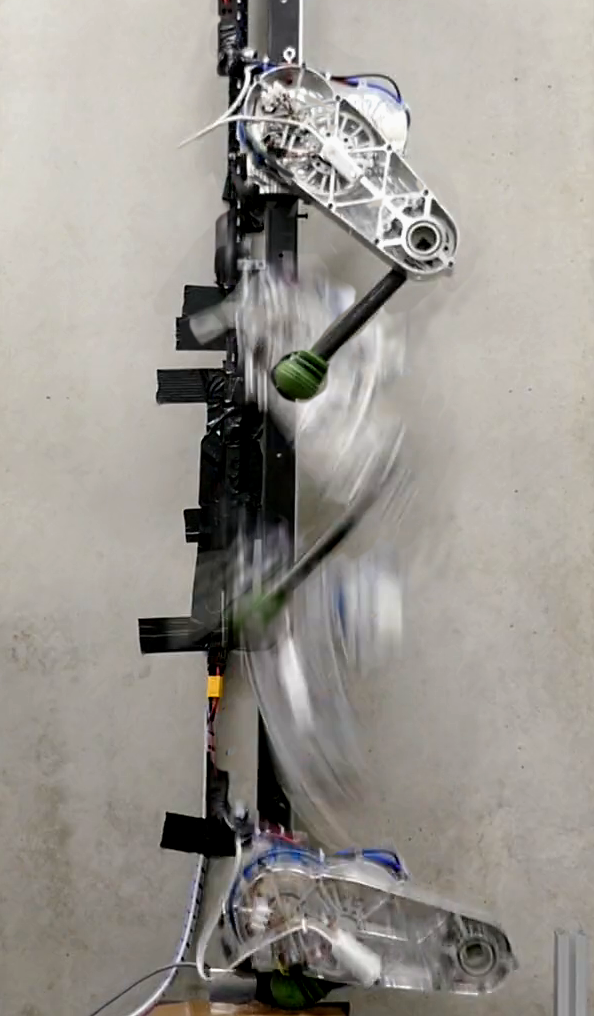}
\caption{Front view of the Capler-Leg prototype mounted on the vertical rail.}\label{fig:caplerleg}
\end{figure}
Legged locomotion is a promising option for mobile systems which navigate through both rough and flat terrains. Many practical solutions were developed but we believe that they are still far from their full potential. Specifically, some of the major problems of existing systems include:
\begin{enumerate}[a.]
\item Prohibitively high cost in designing and manufacturing legged robots.
\item Complicated designs resulting in low degree of robustness.
\item Insufficient autonomy in terms of battery power and energy efficiency. 
\item Insufficient joint acceleration and torque capabilities.
\end{enumerate}

The aforementioned problems are commonly associated with the currently available transmission systems, such as harmonic drives, planetary gears, belt and chain drives, etc. Harmonic drives in particular, add significant weight and complexity to the system due to their sealing requirement, and most importantly are highly inefficient. Commonly reported values for their efficiency are misleading since they are measured at the maximum efficiency condition. The actual efficiency is strongly dependent on temperature, load and speed. 

Planetary gear systems on the other hand add significant weight to the system. They are typically constructed with hardened steel and are usually heavier than the motor. Thus, the overall torque density drops significantly. In addition, planetary gears suffer from backlash and require constant maintenance when they are not properly sealed. Sealing is also costly in terms of weight and complexity. Moreover, a high ratio planetary gear system can be prone to damage during high velocity impacts which are common in legged locomotion.

The proposed alternative is a cable-pulley system based on Capstan drive with a synthetic fiber. Although such solution already exists in the literature~\cite{mazumdar2017synthetic}, we push its capabilities further with a few advances in the design. First, by integrating the pulley to the link itself, our system shows higher power density and lower volume. This made much more dynamics maneuvers possible, as will be presented in Sec.~\ref{sec:exp}. Second, a better choice of material for the cable significantly increased the lifetime of the system, made the design a practical choice for legged systems. Our cable survived after half a million cycles of hopping. Third, we demonstrate extremely high recuperation rate, which is computed using the measurements at the battery output.

There are other notable cable-pulley designs for legged systems. They are usually applied in combination with other types of transmissions such as a planetary gear systems~\cite{dyneemaQuad}, or they become bulky and heavy when using a steel cable~\cite{mabel}. Our cable-pulley system is directly attached to a Permanent-Magnet Synchronous Motor~(PMSM) to maximize the torque transparency and the torque density. The resulting system is very simple thus lowering the likelihood of a malfunction.

The proposed solution also has a few similarities to~\cite{seok2013design}. Both systems are designed to maximize the torque transparency and power density. The proposed solution's advantage is that its transmission is extremely lightweight. It can be machined directly on the links which makes the transmission nearly free of weight. Since there is no direct mechanical friction, it is more efficient as well.

However, there are several challenges in designing a legged robot with a cable-pulley system. Namely, finding a creep resistant cable material, designing a compact and lightweight pulley and making a robust mechanical connection between a pulley and a cable. In this paper we present our solution for cable-driven actuation that addresses all of these challenges. 

For this purpose, we initially developed a single-actuator test-bench for the cable-pulley system and conducted multiple experiments to show the performance and efficiency of the system. Moreover, we designed and constructed a single-limb system Capler-Leg to demonstrate our concept in a simplified planar setting of legged locomotion. The leg weighs approximately \unit[4.0]{kg} and is \unit[57]{cm} in leg length when fully extended. Although it is a relatively small robotic system, it outputs nearly \unit[5,000]{W} of power. We performed a high jump experiment where Capler-Leg hops according to the various commanded heights. The maximum height reached was \unit[1.5]{m} while bearing an added weight of \unit[2.0]{kg}. Since the system is so efficient, almost no energy is lost in the transmission. We achieved a remarkable (\unit[96.5]{\%}) recuperation rate in continuous jumps.

\section{PULLEY TRANSMISSION DESIGN}

In this section, the proposed cable-pulley system is described. The design of such a mechanism comprises three main considerations. The first is the selection of the material for the cable, the second is the design of the pulley and the third the mounting and tensioning of the cable.

\subsection{Cable Material Selection}

\begin{table*}
\begin{tabular}{ |l|c|c|c|c|c|c|c|c|c|}
\hline
   &Strength(MPa)&Stiff.(GPa)&UV res.&Water res.&Creep res.&Density(g/cm$^3$)&D-to-d&Max. Temp($\SI{}{\degreeCelsius}$)&Abr. res.\\
\hline
Stainless Steel 		&689	&200		&1		&1		&1		&7.9	&15$\sim$30	&above 300 & 3\\
\hline
DM20 			& 3500		&50	&1		&1		&1		&0.97		&1$\sim$2		&80 & 1\\
\hline
 Vectran 		& 1100$\sim$3200 	& 50$\sim$100	& 5 &1		&1		&1.4	&2$\sim$3		&140 & 3\\
\hline
  Kevlar 		& 3000 		& 70$\sim$110	&4&5	&3	&1.44	&2$\sim$3		&180 & 5\\
\hline
 Zylon 		& 5800 		& 180	&5&4	&2	&1.56	&2$\sim$3		&650 & 5\\
 \hline
\end{tabular}
 \caption{Cable material property comparison. The values without a unit are only qualitative ratings and interpreted as the following: 1:best, 2:good, 3:average, 4:bad, 5:worst.}
 \label{tab:cablecom}
\end{table*}

Stainless steel is the most popular cable material since it does not creep and termination can be implemented easily using crimping tools. However, this solution is usually inapplicable to legged robots since they require a high D-to-d\footnote{maximum bending diameter to the cable diameter} ratio and have lower tensile strength than fiber cable materials.

Dyneema\textsuperscript{\textregistered} SK70 and SK95 were widely used~\cite{dyneemaHand}, \cite{dyneemaQuad}, and \cite{dyneemaRobot} due to their high tensile strength. However, the main problem with fiber cable systems is that they are prone to creep. Creeping of the rope results in frequent maintenance and can cause unstable behaviors during locomotion. This pitfall makes them impractical in real applications.

However, there is a new promising alternative to the aforementioned materials. Dyneema\textsuperscript{\textregistered} DM20~\cite{dyneemadm20} retains all of their desirable properties while showing high resistance to creep. It is a cable material designed for permanent mooring, which has similar loading conditions (i.e. sinusoidal) to our application. Since the material is relatively new~(it first appeared in the literature in 2012), we could not find any reliable source investigating its applicability in robotics. A few tests on basic material tests can be found in~\cite{kirchhoff2017new}.

In Tab.~\ref{tab:cablecom}, we compare several different cable materials. An important characteristic of cable materials is their resistance to UltraViolate (UV) radiation. Since we are interested in outdoor applications, the cable must be highly resistant to UV radiation. Initially a few non-UV-resistant cables with a jacket were tested but none of the the jackets survived after a few thousand cycles of Cyclic Bending Over Sheave~(CBOS) tests. 

In addition, the abrasion resistance is also important in CBOS applications. Since the pulley material is most likely to be bare or anodized aluminum, the selection of cable material must also take this into account.

Considering all the aspects of the materials, the best material we identified was Dyneema\textsuperscript{\textregistered} DM20. The only apparent downside is that it quickly deteriorates at high temperature which limits the robot's operating maximum operating temperature to well below $\SI{80}{\degreeCelsius}$. The stiffness of DM20 is much lower than that of other fiber materials, but is still comparable to that of Aluminum. Extensive creep tests were  performed~\cite{dyneemadm20} for  over a period of 100 days at $\SI{70}{\degreeCelsius}$ and the authors reported a total creep of less than \unit[0.3]{\%}. This set of experimental condition and duration corresponds to 25 years of operation at room temperature.

\subsection{Pulley Design}

One of the goals of this research is to minimize both the number of parts as well as the overall size of the transmission system. In order to do so, we directly machined the groove features on the rotor hub as well as the output pulley groove on the link. The exact shape is further illustrated in Sec.~\ref{sec:capler} and in Fig.~\ref{fig:testSetup}. Two cables form helices on both pulleys. This ensures that the cable length remains the same when there is no load. The cables on the rotor hub share the helix groove to minimize the size. These two pulleys are tightly placed such that there is almost zero moment produced by the two cables.

\subsection{Cable Mounting and Tensioning}

The key design challenge of the cable-pulley system lies in forming the cable terminations. Since the fiber cables exhibit a much higher tensile strength than metal, it is insufficient to directly press them between metal parts as presented in many literature. Doing so causes loosening and an eventual failure of the terminations.

\begin{figure}
\centering
\includegraphics[width=0.45\textwidth]{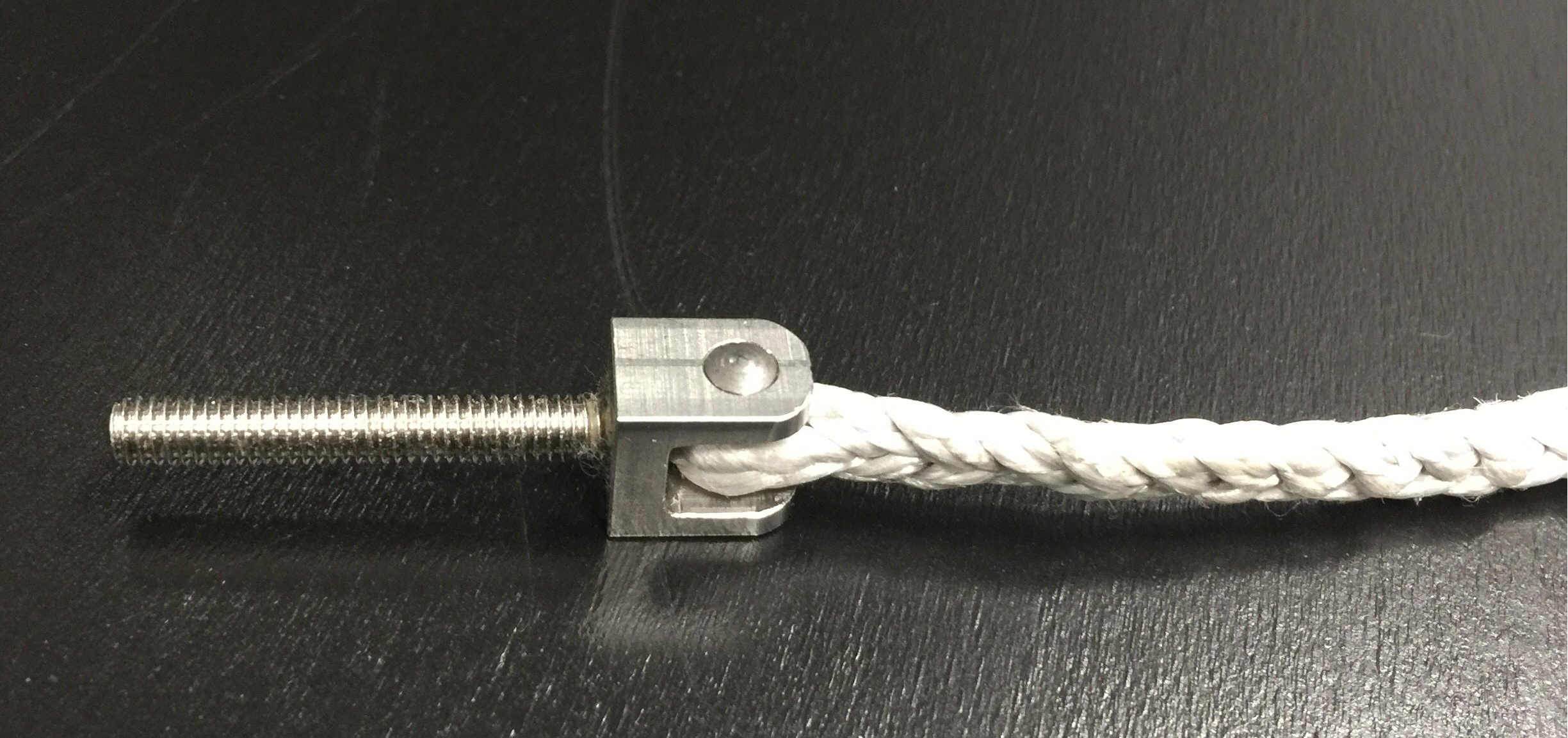}
\caption{Eye-splice termination.}\label{fig:splice}
\end{figure}

In our proposed design, the cable-ends are eye-spliced as shown in Fig.~\ref{fig:splice}. An eye-splice forms a loop at each end such that it can be hung on pins. However, a single eye-splice did not meet our creep resistant requirement in tensile tests when they are repeatedly loaded and relaxed. Therefore, we splice it multiple times in order to create a creep resistant termination. This led to very reliable cable termination. In our experience, common knot terminations are not robust against long periodic loading. Clamping may work by forming multiple loops on metal pieces if it is well-designed. However, the coefficient of friction of Dyneema cables are from 0.05 to 0.08. It is very challenging to make a good mechanism to hold it. Therefore, clamping will result in a bulky and unreliable solution. To the best of our knowledge, there is no test data available regarding creep/loosening for such a type of cable termination. The numerous tests in \cite{dyneemadm20} were performed with eye-splices, and we consider it is to be the most robust and compact option.

\section{EXPERIMENTAL TEST-BENCH}

\begin{figure}
\centering
\includegraphics[width=0.48\textwidth]{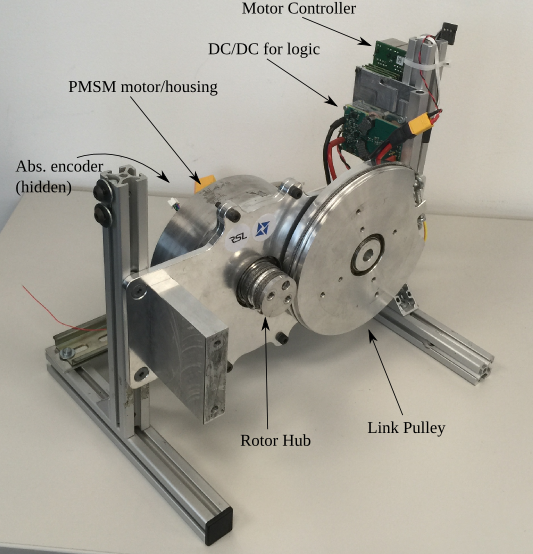}
\caption{Test bench setup.}\label{fig:testSetup}
\end{figure}

To test the cable-pulley design, a test-bench was constructed, as shown in Fig.~\ref{fig:testSetup}. We selected a RoboDrive 115x25~\cite{roboDrive} stator-rotor servo-kit for the actuation, and a single PWB AS50 absolute encoder~\cite{PWBencoder} used simultaneously for both commutation and joint state feedback. Lastly, we used an Elmo Guitar~\footnote{http://www.elmomc.com} to control the current applied to the armatures of the stator.

The low-level motor controller consists of two simple cascaded loops: the outer torque loop running on the host PC computes desired current using the linear relationship $\tau_{m} = K_Ti_{a}^{*}$, and the current loop running on the motor drive performs current tracking to generate the desired motor torque. Since the cable flattens out, the exact pulley radius cannot be measured. We thus computed the exact current-to-torque mapping using an external torque sensor. However, we found that these values deviated by less than 1$\%$ from the pulley radius ratio multiplied by the torque constant provided by the motor manufacturer.

\begin{figure}
\centering
\includegraphics[width=0.48\textwidth]{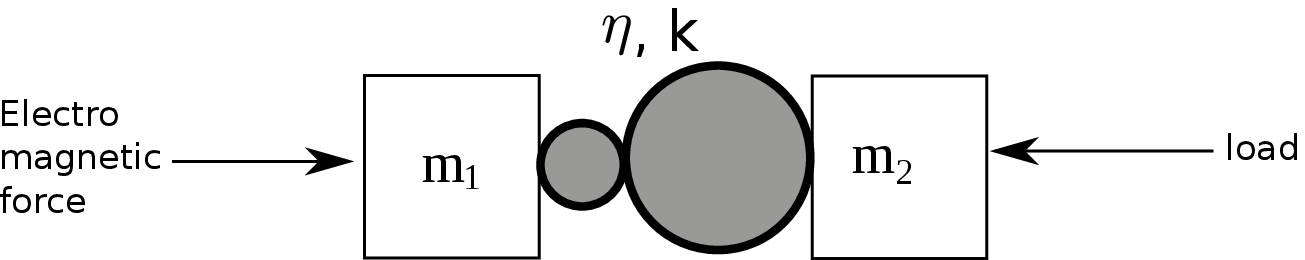}
\caption{Model of the testbench.}\label{fig:testBenchModel}
\end{figure}

We model the rotor and output pulley as one large link as is shown in Fig.~\ref{fig:testBenchModel}. $m_1$, $m_2$ correspond to the rotor and joint inertia respectively.


\begin{figure}
\centering
\includegraphics[width=0.49\textwidth]{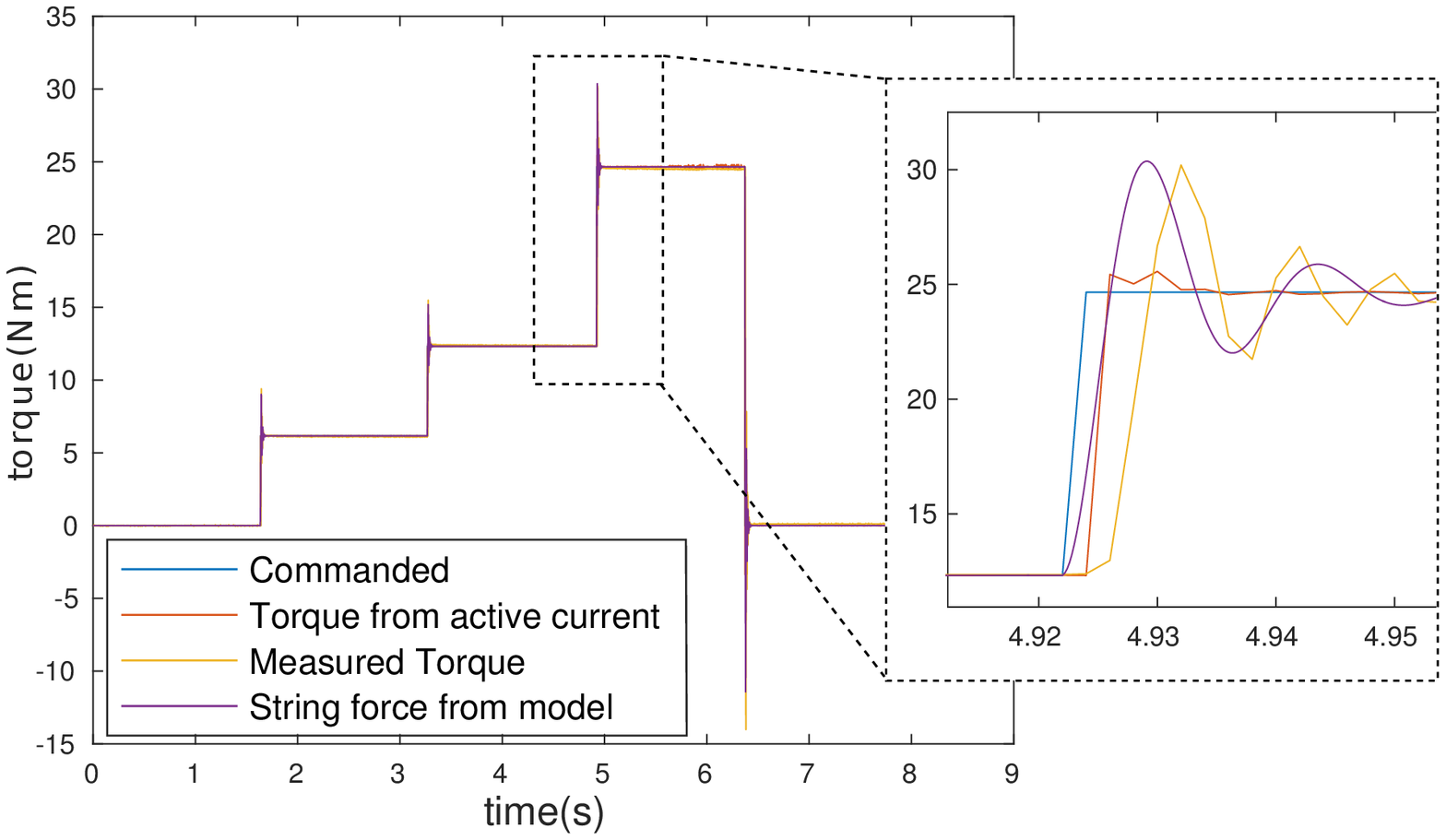}
\caption{Commanded, estimated~(from current), measured and predicted torque~(from model).}\label{fig:step}
\end{figure}

We performed a comparative test between the aforementioned model in simulation and experiments conducted on the test-bench to observe the behavior of the generated torque at the joint with respect to reference desired torques. For this purpose we mounted a torque sensor at the end of the link pulley~($m_2$) and applied step commands with varying magnitudes. For the simulation, we calculated the spring constant according to the Young's modulus given by the manufacturer of the cable and the geometry of the cable-pulley system. Thus, the spring constant was calculated to be \unit[1614]{Nm/rad}, which, is actually in the same order of magnitude as spring constants reported for harmonic drives. The spring damping however was found experimentally as it cannot be easily computed otherwise. Thus, as can be seen in Fig.~\ref{fig:step}, the measured torque from the experiment was found to be very close to the one predicted by the simulation model, with the exception of the time delay which is due to the sampling rate (\unit[500]{Hz}) of the measurement device. The average steady state error was below \unit[0.4]{\%}. This shows that the proposed transmission system is nearly frictionless.

We would like to note that we observed that the effective discrepancy in angular position of the joint was induced by the deflection of the cable. At the maximum torque, this deflection was measured to be around \unit[2]{deg}.

\begin{figure}
\centering
\includegraphics[width=0.49\textwidth]{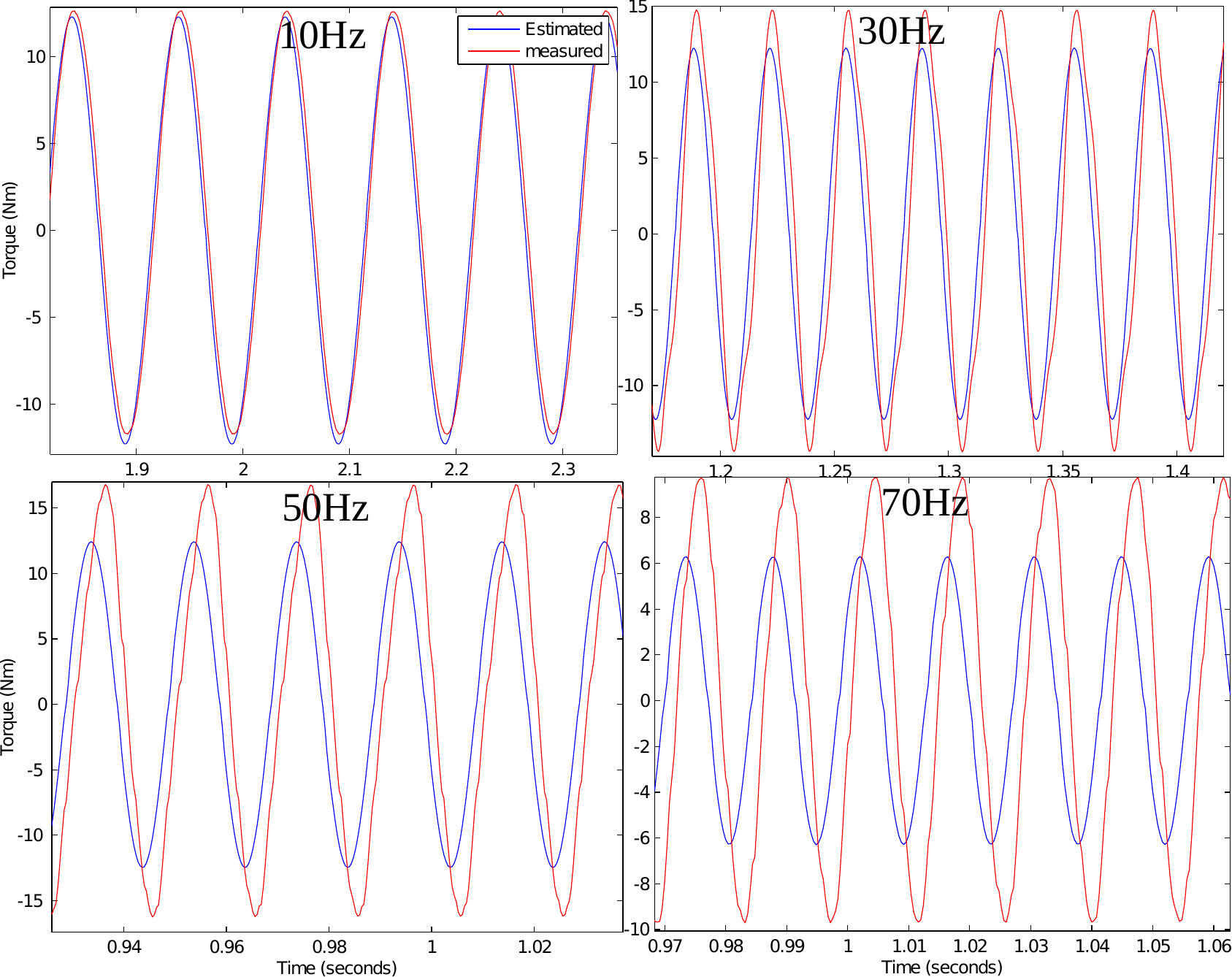}
\caption{Open-loop force tracking behavior at \unit[10, 30, 50, 70]{ Hz} and at \unit[12.5]{Nm} measured at the link end.}\label{fig:SineWavePulley}
\end{figure}

Lastly, we tested the torque tracking at various frequencies and the results are shown in Fig.~\ref{fig:SineWavePulley}. Torque tracking up to \unit[30]{Hz} at \unit[12]{Nm} was reasonably accurate. Note that all the tests are conducted with a simple open-loop current given by the motor drive manufacturer. We did not tune the controller specifically for the experiments. 

\subsection{Comparison to Existing Transmission Systems}
\begin{table*}
\centering
\begin{tabular}{ |l|c|c|c|c|c|c|c|c|c|c|}
\hline
    &Speed              &Backlash                &Torque density          &Torque     & Regeneration& Size                    & Cost \\
    &(rad/s)              &                &(Nm/kg)          &transparency     & efficiency&                     &  \\
\hline
HD &8.5 &2 &110 &5 &5 &1 &4\\
\hline
HD-TS &8.5 &2 &75 &4 &5 &2 &5\\
\hline
SEA &12.0 &2 &40 &2 &5 &3 &5\\
\hline
PG &$>$20 &3 &$\sim$40&2 &2 &3 &3 \\
\hline
CG & 3.5 &2 &50 &5 & 5&3&4\\
\hline
B/C & {N/A}&3 &30 &3 &2 & 5 &2\\
  \hline
DD & $>$100&1 &20 &1 &1&1 &1\\
\hline
CP & $>$40&1 &50 &1&1&3&2\\
 \hline
\end{tabular}
 \caption{This table compares different robotic transmission systems for legged robots. This overview is a qualitative analysis only since they are highly design dependent. The unit-less values are only qualitative ratings and are interpreted as: 1:best, 2:good, 3:average, 4:bad, 5:worst.}
 \label{tab:transcom}
\end{table*}

In order to evaluate transmission systems appropriate for legged robots, we use the following criteria: 1) backlash, 2) torque density, 3) torque transparency 4) regeneration efficiency 5) size and 6) cost. Note that friction, efficiency, and torque transparency are directly related to each other, and thus we consider only torque transparency here.

For the comparison, we examined market leading manufacturers of the transmission systems as well as relevant literature. Since accounting only for the transmission system can constitute an unfair comparison, we consider the whole actuation assembly, assuming that they all use RoboDrive motors as we do. We selected the systems that can produce the same torque as our system. For the systems that cannot produce the required joint speed, we picked the one which matches as closely as possible. The transmission systems we compare in this paper are:
\begin{enumerate}[a.]
\item Harmonic Drives (HD)~\cite{harmonicDrive} 
\item Harmonic Drives with a Torque Sensor (HD-TS)~\cite{ati}
\item Series Elastic Actuators~(SEA)~\cite{anydrive}
\item single-stage Planetary Gear systems~\cite{seokphd} (PG)
\item Cycloidal Gear systems (CG)~\cite{dynamixel}
\item Belt/Chain Drives (B/C D)~\cite{carbonDrive}
\item Direct Drive systems (DD)~\cite{roboDrive}
\item the proposed Cable-Pulley (CP) system
\end{enumerate}
The comparison is shown in in Tab.~\ref{tab:transcom}. For the exact specifications, we used the data from the manufacturers which can be found in the corresponding references. 

The belt drive efficiency from the manufacturer can be misleading since the measurements are measured without pretension. When they are taut, the efficiency is lower than our system since ours consists of an antagonistic pair of cables that produces zero net force on the pulleys. This results in nearly zero friction at the bearings. The backlash is conceptually hard-to-define since many zero-backlash drives show stick-slip behaviors. Although stick-slip behaviors are less severe than backlashes, they are still undesirable for control. Harmonic drives, cycloidal gears, and belt/chain drives all show slip-stick behavior.

For variables which are hard to quantify, we used qualitative ratings from 1 to 5. The weights reflect the additionally required parts, e.g. sealing and/or housing for harmonic drive and sprockets for belt/chain drives.

Since we require minimal or zero backlash, we assume that the drive train is tensioned tightly. A carbon-fiber reinforced belt system~\cite{carbonDrive} was the best belt material we could find but the whole actuation system becomes impractically large due to our high torque requirements. Gear trains can be pressed together to remove the backlash but they wear out much faster and creates high frictional forces. 

Direct drives are nearly ideal but since it cannot produce enough torque, it is not practical at all for medium to large legged robots. It is also very inefficient due to large amount of thermal loses at the motor. For small legged robots, it can also be one possibility~\footnote{https://www.ghostrobotics.io/copy-of-robots}.

\section{Capler-Leg Design}\label{sec:capler}

The results from the test-bench and our thorough analysis led us to design a single leg called \textbf{Ca}ble-\textbf{p}ulley driven \textbf{le}gged \textbf{r}obot~(or Capler for short) Leg. Capler-Leg is a two-joint planar legged robot prototype designed to test the proposed actuation system under conditions closer to that of full legged robots. This section will cover both the mechanical and electrical design of the robot.

\subsection{Capler-Leg}

\begin{figure}
\includegraphics[width=0.45\textwidth]{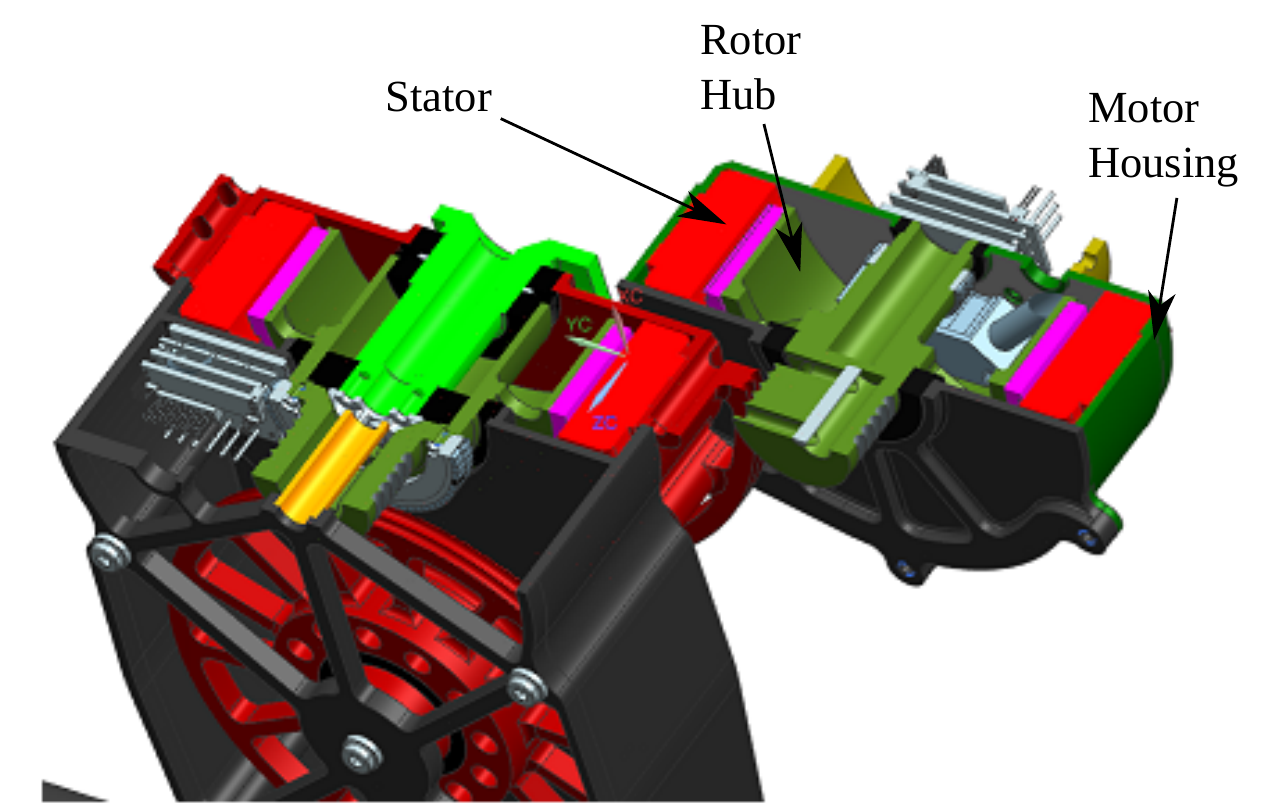}
\caption{Hip cross section view.}\label{fig:legHip}
\end{figure}

\begin{figure}
\includegraphics[width=0.35\textwidth]{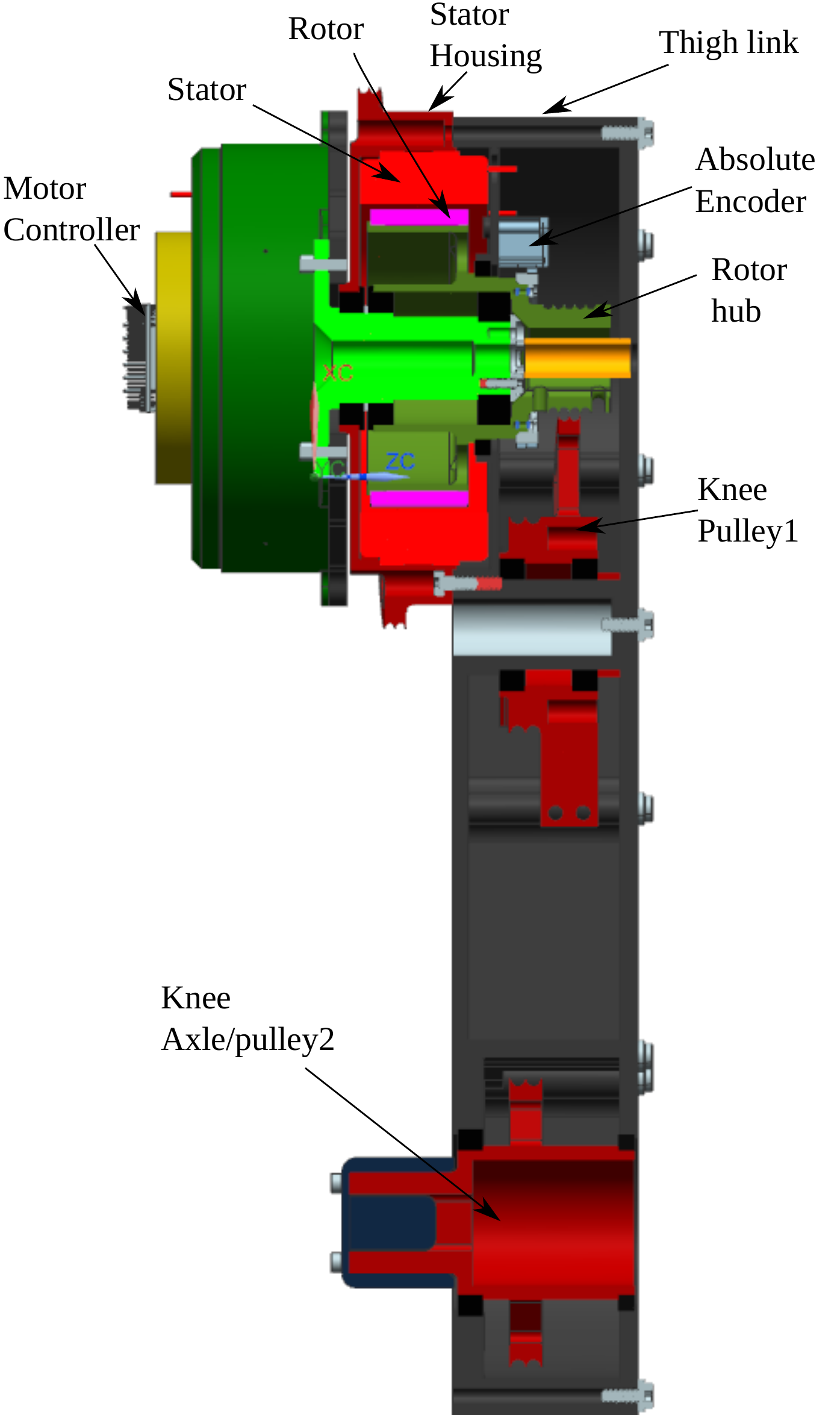}
\caption{Thigh cross section view.}\label{fig:legKnee}
\end{figure}

The design of Capler-Leg is shown in Fig.~\ref{fig:caplerleg} and its interior is shown in the two cross section views in Fig.~\ref{fig:legHip} and Fig.~\ref{fig:legKnee}. The pulley grooves that are directly machined on the motor housing also act as a heat-sink to dissipate heat generated by the knee motor. The thigh is designed to be fully enclosed, due to considerations regarding potential water-proofing assembly in the future. As of yet, we have not performed any sealing test however. The leg is also designed to be rugged so to withstand falls and crashes which are possible since it is capable of highly dynamic maneuvers.

Since the motor is placed at the hip, the power has to be transferred to the knee. In order to do so, two straight pulley grooves are added at the knee pulley and the knee axle. This additional transmission has a gear ratio of 1.33. A thicker cable~(\unit[3]{mm}) was used for this transmission in order to increase the stiffness. The overall gear ratios of the hip and the knee actuator are 5 and 5.33 respectively.

\begin{table}
\centering
\begin{tabular}{ |l|c||l|c|}
\hline
Properties & Values & Properties & Values\\
\hline
Weight & \unit[4.0]{kg} & Max. torque & \unit[70]{Nm}\\
\hline
Link lengths & \unit[24, 25]{cm} & Max. joint speed & \unit[40]{rad/s}\\
\hline
Max. Power & \unit[4200]{W} & Range of motion & \unit[300, 180]{deg}\\
\hline
Leg inertia\footnote{measured at hip when the leg is fully stretched} & \unit[0.062]{kgm$^2$} & torque resolution & \unit[0.07]{Nm}\\
\hline
Static torque error & \unit[0.1]{Nm} & Payload\footnote{maximum load it can bear to stand up from a fully couched configuration} & \unit[20]{kg}\\
\hline
Torque constant\footnote{represents the torque-to-power relationship, for the knee joint} & \unit[4.58]{Nm/$\sqrt{\textrm{W}}$}& & \\
\hline
\end{tabular}
\caption{Specifications of Capler-Leg.}
\label{tab:caplerParam}
\end{table}

The specifications of the leg are summarized in Tab.~\ref{tab:caplerParam}. The resulting output torque is enough for the leg to sustain 7 times of its own weight even while in the most torque-demanding configuration (i.e. fully crouched). The maximum joint speed allows the foot to move at about \unit[20]{m/s} when fully stretched. The extra weights which were placed for the experiments add \unit[2.0]{kg} of mass to the system, making it \unit[6.0]{kg} in total.

\subsection{Experimental Setup}
Actuation at each joint is implemented using the aforementioned pair of RoboDrive~(115X25) and PWB~(AS25 and AS50) encoder. Control of each motor is performed by an embedded motor driver from \textit{Elmo Motion Control}\textsuperscript{\textregistered}\footnote{Elmo Gold-line \textit{Twitter}} with a maximum operating voltage of \unit[100]{V} and peak output current of \unit[55]{A}.

A Lithium-Iron (LiFe) battery provides a low noise (i.e. low voltage-ripple) and high-current power source, and is regulated to a nominal output voltage of \unit[53]{V} with \unit[100]{A} peak current output. 

A linear absolute encoder from SICK\footnote{https://www.sick.com} was used to measure the position and velocity of the base. Measurements of position and velocity are provided of a standard CAN bus at \unit[100]{Hz}, with a resolution of \unit[0.01]{mm}.

\section{LOCOMOTION CONTROL}\label{sec:locomo}

In this section we describe both the control method used to test the operation of Capler-Leg, as well as the respective implementation using our software framework.

\subsection{Stance Phase Controller}
The primary methods we employed for locomotion control are those of a Virtual Model Controller (VMC)~\cite{virtualModelControl} for stance phases (i.e. contact with the ground). We have modeled the rigid body dynamics of the system in generalized coordinates
\begin{equation}
\mathbf{q} = \begin{bmatrix} z_{b} & \phi_{HFE} & \phi_{KFE} \end{bmatrix}^{\top},
\end{equation}
where $z_{b}$ is the vertical position of the base w.r.t. the ground, and $\phi_{HFE}$ and $\phi_{KFE}$ are the joint angles of the Hip and Knee Flexion-Extension joints respectively. Thus the full dynamics of the system is expressed as
\begin{equation}\label{equ:dyn}
\mathbf{M}(\mathbf{q})\ddot{\mathbf{q}} + \mathbf{b}(\mathbf{q},\dot{\mathbf{q}}) + \mathbf{g}(\mathbf{q}) = \, _{I}\mathbf{J}_{c}^{\top}(\mathbf{q})\,_{I}\mathbf{f}_{c} + \mathbf{S}^{\top}\mathbf{\tau},
\end{equation}
where $\mathbf{M}(\mathbf{q})$ is the generalized mass matrix, $\mathbf{b}(\mathbf{q},\dot{\mathbf{q}})$ consists of the generalized Coriolis and centripetal forces and $\mathbf{g}(\mathbf{q})$ is the total generalized forces due to gravity. We assume that the external forces $_{I}\mathbf{f}_{c}$ due to contact with the ground are acting at a pre-specified point at the center of the foot link, which, is modeled as a sphere. This contact force is mapped to generalized forces via the respective Jacobian matrix $_{I}\mathbf{J}_{c}$ corresponding to the absolute linear motion of the foot. The left subscript denotes that the Jacobian matrix and Cartesian forces are expressed in the inertial reference frame \textit{I}. Finally, the selection matrix $\mathbf{S}$ projects the vector of joint torques $\mathbf{\tau}$ to the dimensions of the generalized forces. 

When a contact is detected, we apply VMC to implement a virtual spring acting vertically between the base and the ground. This virtual spring, for a given setting of stiffness $k_{s}$ and damping $d_{s}$, applies a Cartesian force $_{I}\mathbf{f}_{s}$ along the inertial $z$ axis as
\begin{equation}
_{I}\mathbf{f}_{s} = k_{s}\begin{bmatrix}0\\0\\(z^{*}_{b} - z_{b}) \end{bmatrix} + d_{s}\begin{bmatrix}0\\0\\-\dot{z}_{b}\end{bmatrix}.
\end{equation}

By assuming a quasi-static approximation of the full dynamics of the leg, we can neglect the terms involving $\mathbf{M}(\mathbf{q})$ and $\mathbf{b}(\mathbf{q},\dot{\mathbf{q}})$. The contact force becomes $_{I}\mathbf{f}_{c} = [0,0,-mg]^{\top}$, where $m$ is the total mass of the robot and $g$ is the gravitational acceleration. Equation (\ref{equ:dyn}) under our assumption and the virtual force becomes
\begin{equation}
\mathbf{\tau}^{*}_{VMC} = \mathbf{S}(-_{I}\mathbf{J}^{\top}_{c}(\mathbf{q})\,_{I}\mathbf{f}_{c} + _{I}\mathbf{J}_{b\,s}^{\top}(\mathbf{q}) \, _{I}\mathbf{f}_{s} -\mathbf{g}(\mathbf{q})),
\end{equation}
where $_{I}\mathbf{J}^{\top}_{b\,s}$ is the positional Jacobian matrix mapping generalized velocities to the base velocity.

\subsection{Flight Phase Controller}
While the foot is flight phase, i.e. the contact with the ground is open, we apply inverse dynamic control in joint-space with a desired acceleration $\ddot{\mathbf{q}}^{*}_{j}$ specified using a linear PID law on desired joint position $\mathbf{q}^{*}$ and zero joint velocity
\begin{equation}
\ddot{\mathbf{q}}^{*} = K_{P}\left(\mathbf{q}^{*} - \mathbf{q}\right) + K_{I}\int^{t_{f}}_{t_{0}}{\left(\mathbf{q}^{*} - \mathbf{q}\right)dt} - K_{D}\dot{\mathbf{q}}.
\end{equation}
$K_{P}$, $K_{I}$ and $K_{D}$ are specified as diagonal matrices to allow for easy tuning of these gains. The hat over the modeled dynamic quantities denotes that these are estimates given the current measurements and implies a degree of modeling uncertainty. The final torque command resulting from the inverse dynamic controller is computed as
\begin{equation}
\mathbf{\tau}^{*}_{IDC} = \, \mathbf{S} \left( \hat{\mathbf{M}}(\mathbf{q})\ddot{\mathbf{q}}^{*} + \hat{\mathbf{b}}(\mathbf{q},\mathbf{\dot{q}}) + \hat{\mathbf{g}}(\mathbf{q}) \right).
\end{equation}

In order to perform repetitive hopping motions, we selected a relatively low spring damping and tune it so that when excited sufficiently by an external disturbance, the VMC virtual spring oscillates mildly, causing the leg to eventually jump upwards. Furthermore, we enforce a switching to the flight-phase controller when the leg extends beyond \unit[0.4]{m}. These heuristics allow Capler-Leg to passively perform continuous hopping, without needing any sort of motion planning. We found that, in practice, this method proved more than sufficient for our purposes and actually quite robust.  

\section{EXPERIMENTS}\label{sec:exp}
We performed two sets of experiments which we present in this section: a high jump test which tests the capabilities of Capler-Leg, and a longevity test which tests its endurance under realistic loading conditions. The following quantities were used for our analysis:

\begin{enumerate}
\item Power consumption of the battery: It is an average battery power output and is computed as $P_b = V_bI_b$, where $V_b$ is the battery voltage, and $I_b$ is the battery current.
\item Thermal losses of the motors: This is the resistive losses of the motors and is computed as $P_J = R_mI_m^2$, where $R_m$ is the single-phase equivalent motor resistance and $I_m$ is the single-phase equivalent motor current.
\item Motor controller power consumption ($P_e$): It is the average power required to run the motor controllers at idle state, mainly used to power the logic circuits and the communication layer of the motor controllers.
\item Mechanical power: This is a mechanical power from the motors, and is computed as $P_{mech} =\omega_m\tau_m$, where $\omega_m$ is the rotor speed and $\tau_m$ is the motor torque.
\item Mechanical losses: This is the average power lost due to various friction sources such as the guide rail, the transmission and the foot impacts. It is computed as an average of mechanical power. Note that this is accounting for the recuperated energy so only the lost energy is computed.
\end{enumerate}

\subsection{Continuous High Jump}
To test the performance limits of Capler-Leg, we conducted a high-jump experiment, whereby Capler-Leg is mounted on a vertical rail of \unit[1.5]{m} height, and, using the simple locomotion controller described in Sec.~\ref{sec:locomo}, jumped multiple times with various target heights. During these high-jump heights, the base was often crashing onto the foot while landing and high mechanical losses arose. However, we present the power computation for completeness.

\begin{figure}[h]
\centering 
\includegraphics[width=0.48\textwidth]{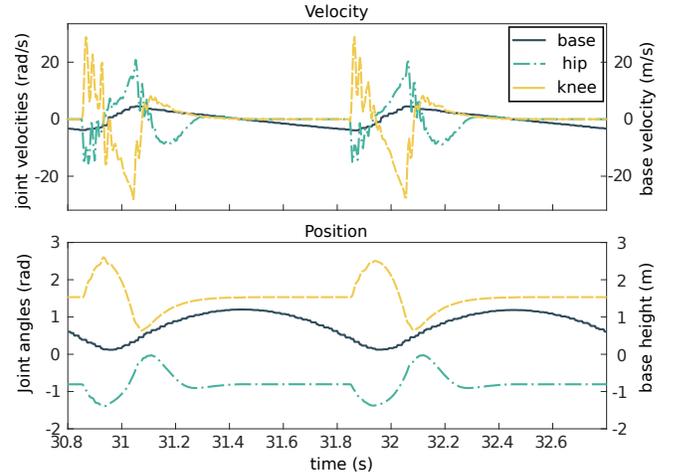}
\caption{Generalized coordinate and velocity trajectories during the high-jump test.}\label{fig:hjt}
\end{figure}

Figure~\ref{fig:caplerleg} show the snapshots of the leg while jumping. The measured states of the leg is shown in Fig~\ref{fig:hjt}. The maximum torque and the joint velocity observed during the experiment were \unit[52.0]{Nm} and \unit[35.1]{rad/s} respectively. The maximum achieved height was \unit[1.22]{m}. 

\begin{figure}
\centering 
\includegraphics[width=0.48\textwidth]{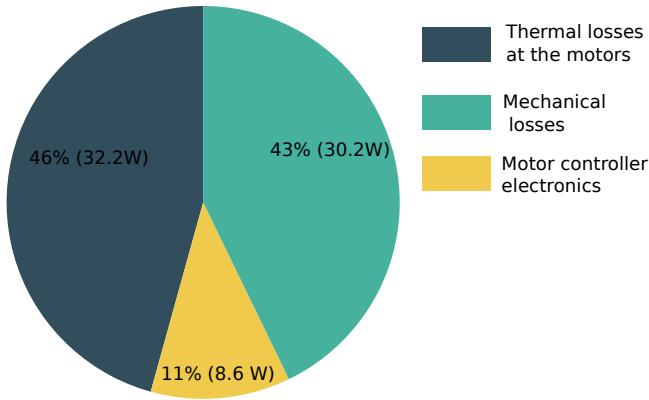}
\caption{Power losses during the high-jump test is shown. The total power loss, \unit[70.20]{W}, was measured at the battery.}\label{fig:pieChartHJ}
\end{figure}

The power related statistics from the experiments are shown in Fig.~\ref{fig:pieChartHJ}. The consumed power was \unit[70.20]{W} including the \unit[9.0]{W} from the electronics. At equilibrium, the temperature of the motor was found to be \unit[32.8]{$^\circ$C} at most.

\begin{figure}
\centering 
\includegraphics[width=0.48\textwidth]{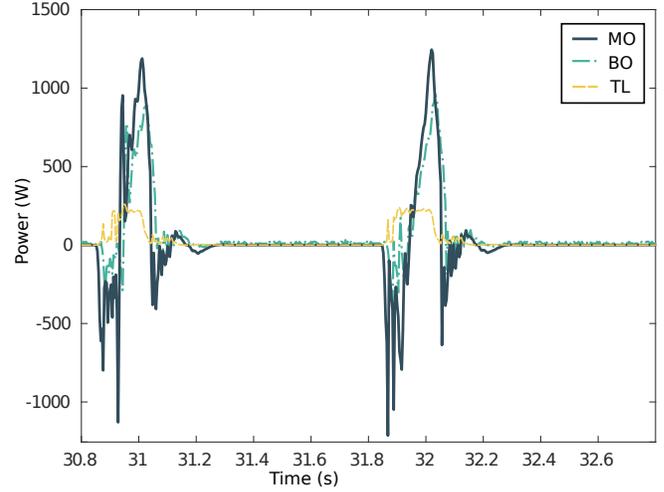}
\caption{Power and height trajectories during high-jump test. The acronyms stands for, BO: Battery Output, MO: Mechanical Output of the motors, TL: Thermal losses of the motors. Note that power consumption can be greater than the power supply since there is a capacitor regulating the bus voltage.}\label{fig:hjp}
\end{figure}

\subsection{Longevity Test}
The longevity experiments were mainly designed to test the endurance of the cables over a long period of operation, but also provided means to evaluate the system under realistic operating conditions. We require that the cable should last at least half a million cycles before needing re-tensioning or replacement. Assuming that the robot moves \unit[40]{cm} in every step, this test corresponds to \unit[200]{km} of total travel distance. 

\begin{figure}[h]
\centering 
\includegraphics[width=0.48\textwidth]{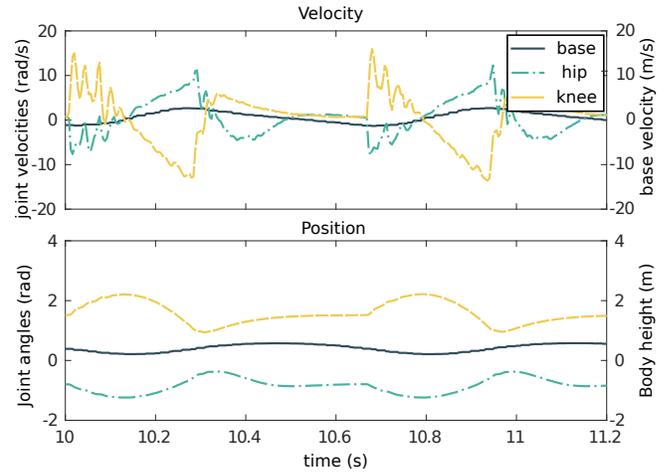}
\caption{Generalized coordinate and velocity trajectories during the longevity test.}\label{fig:lt}
\end{figure}

\begin{figure}
\centering 
\includegraphics[width=0.48\textwidth]{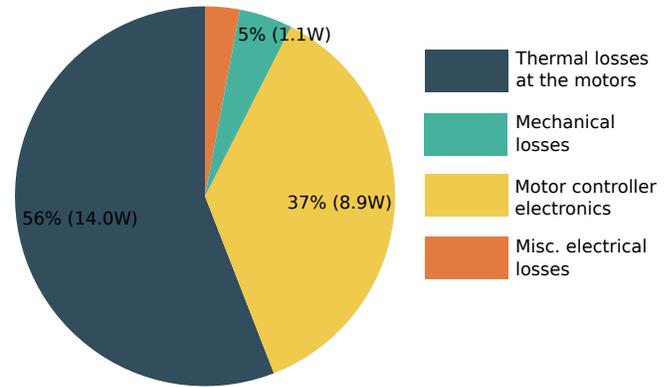}
\caption{Power losses during the longevity test. The total power loss, \unit[25.12]{W}, was measured at the battery. The acronyms stands for, BO: Battery Output, MO: Mechanical Output of the motors, TL: Thermal losses of the motors. }\label{fig:pieChartL}
\end{figure}

During these experiments, Capler-Leg continuously hopped with an average base height change of \unit[35]{cm} and a foot clearance of \unit[20]{cm}. This simulates the loading condition of moderate running on a flat terrain. The power and height trajectories for one period of jump are shown in Fig.~\ref{fig:lt}. Some statistics on different power values are shown in Fig.~\ref{fig:pieChartL}. During hopping, the average positive work of the motors was \unit[28.83]{W} and the average negative work of the motors was \unit[27.85]{W}. The mechanical power output was measured at the motor but the negative work was computed as
\begin{equation}
	P_{recup} = P_j + P_{mech} + P_e - P_b.
\end{equation}
Therefore, this is the true value for the energy harvested at the battery due to negative mechanical work. This means that \unit[96.5]{\%} of the mechanical energy was recuperated at the motors. The efficiency in terms of electrical to mechanical conversion could not be measured since there are many electrical energy storages (capacitance and inductance) and the recuperation energy was also present. Only \unit[4.5]{\%} of the total energy consumption was contributed by the mechanical losses such as transmission losses, impact losses and friction losses. \unit[14.04]{W} of power was lost at the motors due to joule heating. Although it contributes more than \unit[50]{\%} of the power consumption, it was not enough to increase the surface temperature of the motor by \unit[3]{$^\circ$C} at equilibrium. The motor controllers consuming significant amount of power as well mainly due to the fast communication protocol (\textit{EtherCAT}).

\begin{figure}
\centering 
\includegraphics[width=0.48\textwidth]{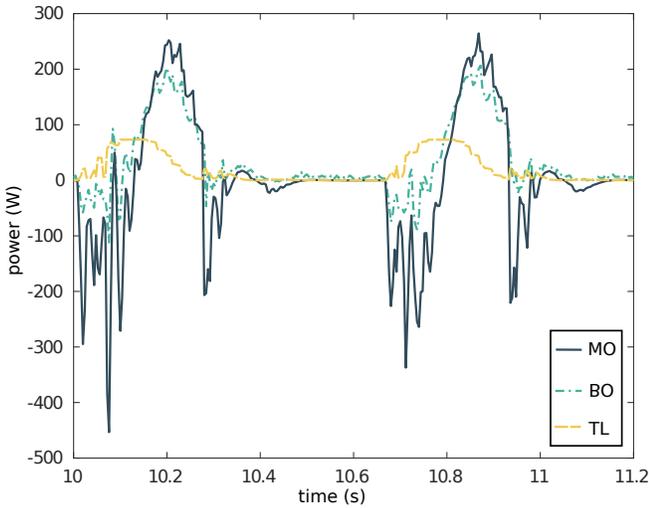}
\caption{Power and height trajectories during the longevity test. The acronyms stands for, BO: Battery Output, MO: Mechanical Output of the motors, TL: Thermal losses of the motors. Note that power consumption can be greater than the power supply since there is a capacitor regulating the bus voltage.}\label{fig:lp}
\end{figure}

The total test lasted about \unit[42]{hours}. The jumping behavior did not change throughout the experiment which is indicative that there is no significant degradations in the cables. We also visually inspected the cables for any possible damage or creep and the cables were visually intact.

\section{CONCLUSIONS}
This paper presented a novel cable-pulley system and the derived single legged robotic limb, Capler-Leg. The proposed system has a gear ratio of 5 which is much lower than the transmissions in most legged robots. However, in conjunction with a high gap-radius motor, we show that the combined system is not just efficient, but also extremely powerful. We conducted several tests for torque tracking and it was extremely accurate both for static and dynamic cases. We presented a high-jump test of \unit[1.2]{m} only using \unit[80]{\%} of the torque capacity with an extra weight which is \unit[50]{\%} of the robot's own weight. We also showed longevity test of half a million cycles and proved that the cable is durable in practical scenarios. During the longevity test, Capler-Leg recuperated \unit[96.5]{\%} of the total mechanical power in average. This shows that the transmission has extremely low friction. Therefore, we conclude that the proposed solution is very well-suited for legged robotics. 


\bibliographystyle{IEEEtran}
\bibliography{references}

\begin{thebibliography}{10}

\bibitem{mazumdar2017synthetic}
A.~Mazumdar, S.~J. Spencer, C.~Hobart, J.~Dabling, T.~Blada, K.~Dullea,
  M.~Kuehl, and S.~P. Buerger, ``Synthetic fiber capstan drives for highly
  efficient, torque controlled, robotic applications,'' \emph{IEEE Robotics and
  Automation Letters}, vol.~2, no.~2, pp. 554--561, 2017.

\bibitem{dyneemaQuad}
S.~Kitano, S.~Hirose, G.~Endo, and E.~F. Fukushima, ``Development of
  lightweight sprawling-type quadruped robot titan-xiii and its dynamic
  walking,'' in \emph{2013 IEEE/RSJ International Conference on Intelligent
  Robots and Systems}.\hskip 1em plus 0.5em minus 0.4em\relax IEEE, 2013, pp.
  6025--6030.

\bibitem{mabel}
J.~Grizzle, J.~Hurst, B.~Morris, H.-W. Park, and K.~Sreenath, ``Mabel, a new
  robotic bipedal walker and runner,'' in \emph{2009 American Control
  Conference}.\hskip 1em plus 0.5em minus 0.4em\relax IEEE, 2009, pp.
  2030--2036.

\bibitem{seok2013design}
S.~Seok, A.~Wang, M.~Y. Chuah, D.~Otten, J.~Lang, and S.~Kim, ``Design
  principles for highly efficient quadrupeds and implementation on the mit
  cheetah robot,'' in \emph{Robotics and Automation (ICRA), 2013 IEEE
  International Conference on}.\hskip 1em plus 0.5em minus 0.4em\relax IEEE,
  2013, pp. 3307--3312.

\bibitem{dyneemaHand}
W.~Friedl, M.~Chalon, J.~Reinecke, and M.~Grebenstein, ``Frcef: The new
  friction reduced and coupling enhanced finger for the awiwi hand,'' in
  \emph{Humanoid Robots (Humanoids), 2015 IEEE-RAS 15th International
  Conference on}.\hskip 1em plus 0.5em minus 0.4em\relax IEEE, 2015, pp.
  140--147.

\bibitem{dyneemaRobot}
J.-B. Izard, M.~Gouttefarde, M.~Michelin, O.~Tempier, and C.~Baradat, ``A
  reconfigurable robot for cable-driven parallel robotic research and
  industrial scenario proofing,'' in \emph{Cable-Driven Parallel Robots}.\hskip
  1em plus 0.5em minus 0.4em\relax Springer, 2013, pp. 135--148.

\bibitem{dyneemadm20}
M.~Vlasblom, B.~Fronzaglia, S.~Leite, P.~Davies, and J.~Boesten, ``Development
  of hmpe fiber for deepwater permanent mooring applications,'' \emph{Work},
  vol.~3, no. 3.5, pp. 3--6, 2012.

\bibitem{kirchhoff2017new}
J.~Kirchhoff and O.~von Stryk, ``New insights in synthetic fiber rope
  elongation and its detection for ultra lightweight tendon driven series
  elastic robots,'' in \emph{Advanced Intelligent Mechatronics (AIM), 2017 IEEE
  International Conference on}.\hskip 1em plus 0.5em minus 0.4em\relax IEEE,
  2017, pp. 64--69.

\bibitem{roboDrive}
``robodrive,'' \url{http://www.robodrive.com/}, accessed: 2016-09-10.

\bibitem{PWBencoder}
``Pwb encoder,'' \url{http://www.pwb-encoders.com}, accessed: 2016-09-10.

\bibitem{harmonicDrive}
``Harmonic drive,''
  \url{http://harmonicdrive.de/mage/media/catalog/category/2014_11_ED_1019646_CSD_2UH_2UF_2.pdf},
  accessed: 2016-09-10.

\bibitem{ati}
``Ati industrial automation,'' \url{http://www.ati-ia.com/}, accessed:
  2016-09-10.

\bibitem{anydrive}
M.~Hutter, C.~Gehring, D.~Jud, A.~Lauber, C.~D. Bellicoso, V.~Tsounis,
  J.~Hwangbo, K.~Bodie, P.~Fankhauser, M.~Bloesch \emph{et~al.}, ``Anymal-a
  highly mobile and dynamic quadrupedal robot,'' in \emph{Intelligent Robots
  and Systems (IROS), 2016 IEEE/RSJ International Conference on}.\hskip 1em
  plus 0.5em minus 0.4em\relax IEEE, 2016, pp. 38--44.

\bibitem{seokphd}
S.~Seok, ``Highly parallelized control programming methodologies using
  multicore cpu and fpga for highly dynamic multi-dof mobile robots, applied to
  the mit cheetah,'' Ph.D. dissertation, Massachusetts Institute of Technology,
  2014.

\bibitem{dynamixel}
``Robotis,'' \url{http://www.robotis.com/xe/dynamixel_en}, accessed:
  2016-09-10.

\bibitem{carbonDrive}
``gates, carbon drive,'' \url{http://www.gates.com/}, accessed: 2016-09-10.

\bibitem{virtualModelControl}
J.~Pratt, C.-M. Chew, A.~Torres, P.~Dilworth, and G.~Pratt, ``Virtual model
  control: An intuitive approach for bipedal locomotion,'' \emph{The
  International Journal of Robotics Research}, vol.~20, no.~2, pp. 129--143,
  2001.

\end{thebibliography}

\end{document}